\documentclass[sigconf,screen]{acmart}

\setcopyright{cc}
\setcctype{by}
\acmDOI{10.1145/3757279.3785641}
\acmYear{2026}
\copyrightyear{2026}
\acmISBN{979-8-4007-2128-1/2026/03}
\acmConference[HRI '26]{Proceedings of the 21st ACM/IEEE International Conference on Human-Robot Interaction}{March 16--19, 2026}{Edinburgh, Scotland, UK}
\acmBooktitle{Proceedings of the 21st ACM/IEEE International Conference on Human-Robot Interaction (HRI '26), March 16--19, 2026, Edinburgh, Scotland, UK}
\received{2025-09-30}
\received[accepted]{2025-12-01}

\usepackage{booktabs}
\usepackage{afterpage}
\usepackage{tabularx}
\usepackage{array}
\usepackage{siunitx}
\usepackage{algorithm}
\usepackage[noend]{algpseudocode}
\usepackage{microtype}
\usepackage[nameinlink,noabbrev,capitalize]{cleveref}
\usepackage{pgfplots}
\pgfplotsset{compat=1.18}

\usepackage{placeins}
\usepackage{enumitem}
\usepackage{textcomp}
\usepackage{adjustbox}

\usepackage{caption}
\begin{document}

\title[SCOPE: Real-Time Natural Language Camera Agent at the Edge]%
      {SCOPE: Real-Time Natural Language Camera Agent at the Edge}
\subtitle{A Sim-to-Real Benchmark and Analysis of Open-Source Vision and Language Agents for PTZ Camera Tasks}

\author{Nikolaj Hindsbo}
\orcid{0009-0000-5472-2461}
\email{nh@armada.ai}
\affiliation{%
  \institution{Armada AI}
  \city{Bellevue}
  \state{Washington}
  \country{USA}
}

\author{Sina Ehsani}
\email{se@armada.ai}
\orcid{0000-0002-6009-7612}
\affiliation{%
  \institution{Armada AI}
  \city{Bellevue}
  \state{Washington}
  \country{USA}
}

\author{Pragyana Mishra}
\orcid{0000-0002-6111-797X}
\email{pm@armada.ai}
\affiliation{%
  \institution{Armada AI}
  \city{Bellevue}
  \state{Washington}
  \country{USA}
}

\begin{abstract}
Deploying language-driven agents in robotics requires evaluations that reflect real-world task demands: natural-language instructions with reproducible outcomes. Such agents must connect language models to callable perception and control tools, and be assessed using deployment-critical metrics including latency, accuracy, and error modes. We present SCOPE (Simulation and Camera Operations for Perception and Evaluation), a modular agent for natural-language, open-vocabulary pan--tilt--zoom (PTZ) camera control and visual scene understanding, designed explicitly for edge deployment. SCOPE operates both in a Blender-based simulation environment and on a physical PTZ camera, executing all perception, planning, and control locally at the deployment site using edge-accessible compute.

We introduce a \href{https://github.com/HindsboNikolaj/SCOPE}{Blender-based agent environment} that exposes realistic PTZ control affordances and enables reproducible, language-driven tasks aligned with real-world camera operation. Using this environment, we release a 536-task benchmark spanning QA, single- and multi-step commands, counting, spatial reasoning, descriptions, and optical character recognition. Execution traces are combined with an LM-as-Judge to evaluate latency, accuracy, and error modes.

We evaluate 19 planner--perception model combinations pairing Qwen3 small language models (SLMs) with Moondream and Qwen vision--language models (VLMs) on our benchmark. Stronger SLMs substantially reduce hallucinations and improve tool routing, leading to more reliable closed-loop behavior. Once a sufficiently capable SLM is used, perception becomes the dominant performance bottleneck. Architectural choices further shape deployability: Mixture-of-Experts models on both the planning and perception side consistently match or exceed dense alternatives while operating at latencies and memory footprints comparable to much smaller networks. Quantization provides additional efficiency gains with minimal accuracy degradation. Together, these results identify a practical, sim-to-real--validated design point for real-time, edge-feasible language-driven PTZ control.
\end{abstract}



\maketitle

\begin{figure*}[t]
  \centering
  \includegraphics[width=.90\textwidth]{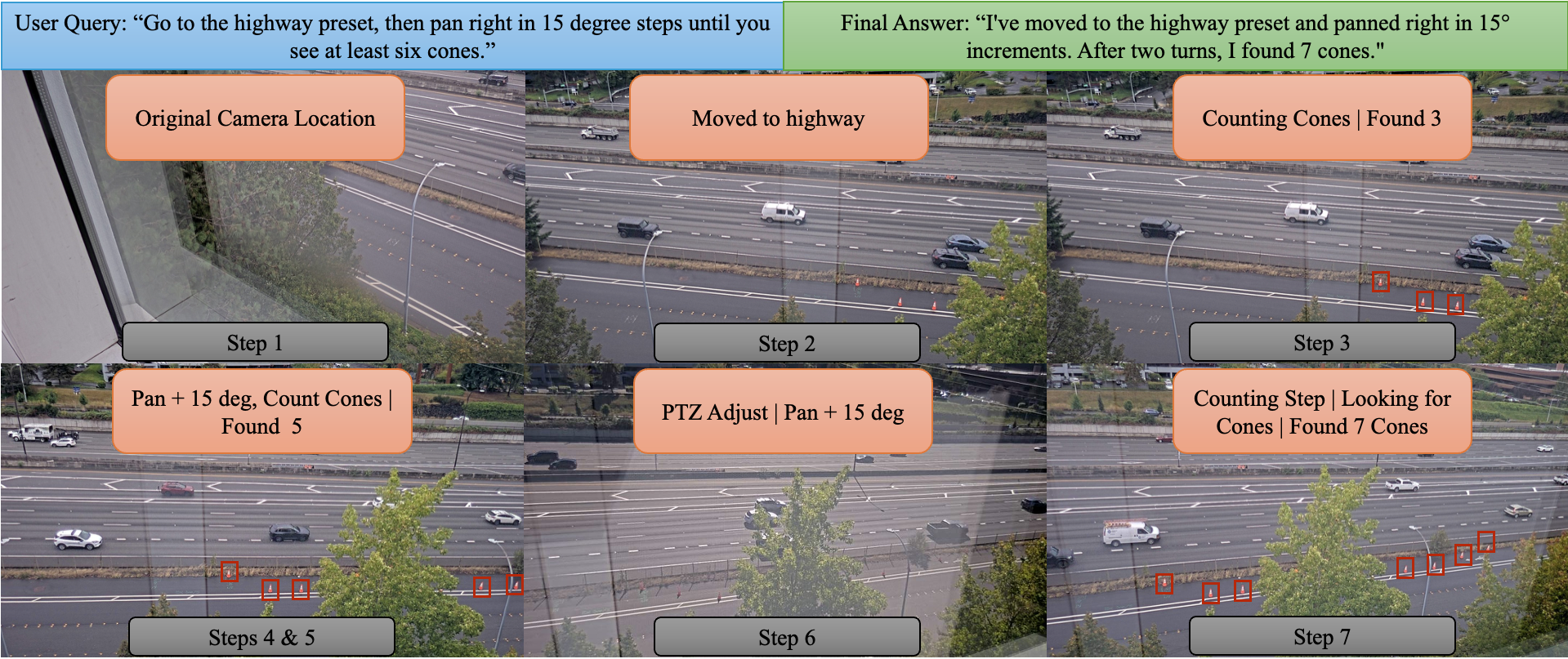}
  \caption{Example of SCOPE executing a multi-step language-guided PTZ task on our physical camera. The user requests: ``Go to the highway preset, then pan right in $15^\circ$ steps until you see at least six cones.'' The agent moves to the preset, performs incremental panning, and invokes the VLM to count cones after each step. For visualization clarity in the paper, cones are shown with bounding boxes rather than the point-based indicators returned by the system at runtime. Step-wise annotations summarize camera actions and intermediate counts, and the final response aggregates these observations, confirming seven cones.}
  \Description{Highway scene viewed by a PTZ camera; cones boxed; banners show actions and counts across pans; final count equals seven.}
  \label{fig:highway-example}
\end{figure*}

\section{Introduction}

Operating a pan--tilt--zoom (PTZ) camera from natural language sits at the intersection of perception, planning, and precise actuation under strict latency and compute constraints. A seemingly simple instruction such as ``go to the highway preset, then pan right in $15^\circ$ steps until you see at least six cones'' expands into a non-trivial feedback loop that couples perception, reasoning, and motion. Figure~\ref{fig:highway-example} shows this process unfolding in a real end-to-end deployment of our language-driven PTZ agent. Executing such tasks autonomously on edge hardware is challenging because success depends not on isolated perception or reasoning, but on reliable coordination of language, vision, and actuation under tight memory and latency budgets.

Recent benchmarks demonstrate rapid progress in general reasoning~\cite{hendrycks2020measuring}, multimodal understanding~\cite{yue2024mmmu}, and tool invocation~\cite{patil2025bfcl}. However, these evaluations remain indirect for embodied, language-guided control. Most do not measure closed-loop behavior, struggle to make sim-to-real claims, and do not capture deployment-critical constraints such as latency, cost, and error modes on edge hardware. As a result, it remains difficult to predict how a language-driven agent will behave—and fail—when deployed in real robotic control loops.

To address this gap, we present Simulation and Camera Operations for Perception and Evaluation (SCOPE), a Blender-based framework and benchmark for executable, closed-loop evaluation of language-driven PTZ camera agents. By exposing an identical PTZ action space and observation loop in both simulation and on a physical camera, SCOPE enables faithful sim-to-real transfer and evaluation for PTZ agents.
Our contributions are as follows:
\begin{itemize}
  \item We introduce SCOPE, a modular, language-driven PTZ camera agent that operates both in simulation and on a physical camera, enabling executable and closed-loop control under real deployment constraints.

  \item We present a Blender-based simulation framework that mirrors the PTZ action space and observation loop seen by the small language model (SLM) planner on real hardware, enabling faithful sim-to-real evaluation of language-driven camera agents.

  \item We release a 536-task benchmark of natural-language PTZ tasks instantiated within a Blender-based simulation environment. The benchmark requires active viewpoint control and closed-loop interaction, reflecting real camera operation and exposing failure modes not observable in static image benchmarks.

  \item We conduct a systematic evaluation of 19 model combinations pairing planners and perception backbones. Using this benchmark, we measure accuracy, latency, and error modes to identify architectural and quantization choices that enable real-time, edge-feasible PTZ deployment.
\end{itemize}
\section{Related Work}


In robotics, language models are increasingly used as planners that call grounded skills rather than as low-level controllers. \emph{PaLM-SayCan} grounded open-ended instructions in robot affordances and value functions, showing that language plans become useful when constrained by what the robot can actually do \cite{ahn2022can}. \emph{Code as Policies} showed that language models can produce code to control robots on-the-fly, demonstrating generalization while highlighting the need for tight output constraints to avoid execution errors \cite{liang2022code}. Conversational prompting for robotics with capability schemas has also been examined (e.g., ChatGPT for robotics), emphasizing structured formats to reduce invalid commands \cite{vemprala2024chatgpt}. These works underscore the potential of SLMs as high-level planners but also the need to ground decisions in actual actions to avoid hallucinations or infeasible plans.

While Large Language Model (LLM) planning is now well established, most work on language-driven active perception has focused on tasks involving mobile agents. Foundational benchmarks like Vision-and-Language Navigation \citep{anderson2018vision, shridhar2020alfred} and Embodied Question Answering  \citep{das2018embodied} require an agent to physically navigate complex 3D environments to find objects or answer questions. Although this work is highly relevant, the focus on mobile navigation means these benchmarks do not address the specific challenges of controlling a static PTZ camera. Furthermore, their evaluation criteria are not centered on the sim-to-real parity of tool-calling agents or the deployment-critical metrics (e.g., latency, cost) essential for edge hardware, leaving this distinct problem domain largely under-explored.


This opportunity is unlocked by modern vision-language models (VLMs) (e.g., Qwen2.5-VL~\cite{bai2025qwen2}, Moondream~\cite{moondream2}, and Molmo~\cite{deitke2025molmo}), which have become increasingly useful for robotics. Beyond free-form captions, many now emit structured outputs—counts, optical character recognition (OCR), strings, attributes, and spatial primitives (points/boxes)—often encoded as JSON, making them directly consumable by planners. Grounded supervision with points/boxes and large open synthetic datasets (e.g., PixMo~\cite{deitke2025molmo}) has reduced counting and commonsense biases and improved robustness against memorization (e.g., stock footage of clock faces or animals with extra limbs/parts that break dataset priors~\cite{moondream2}). 

Lightweight VLMs in our study also reliably emitted JSON-formatted outputs without parsing issues. Moondream, in particular, is designed for efficient deployment on commercial hardware: it runs on an NVIDIA RTX 3090 with a memory footprint of 2.45\,GB and achieves up to 184 tokens/s throughput~\cite{moondream2}. We validated this efficiency in our own evaluation; when hosted locally on an A100-80GB GPU, all lightweight VLMs produced consistent query and counting responses within 200–500\,ms, supporting their suitability for real-time control loops. For clarity, we define “real-time” in this context as sustaining an approximately 2\,Hz control-loop frequency ($\leq$500\,ms per inference), which is sufficient for PTZ camera scanning and framing tasks. In contrast, our Blender-based simulation evaluations relied on a shared cloud cluster, where network and scheduling overheads led to longer response times.



Prior to SCOPE, we developed OPUS~\citep{buynitsky2025camera}, which demonstrated a few-shot, tool-calling pipeline for language-driven PTZ control. While functional, OPUS exhibited several practical limitations that constrained extensibility and deployment. In particular, introducing new tools required task-specific data generation and fine-tuning. Furthermore, perception was handled through You Only Look Once (YOLO)~\cite{redmon2016yolo}–style detectors, which limited the range of visual queries supported.

\section{System Architecture}\label{architecture}

\begin{figure}[tb!]
  \centering
  \includegraphics[width=\linewidth]{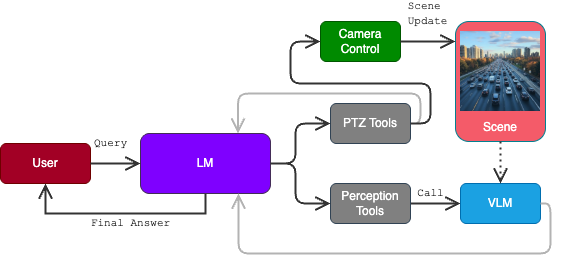}
  \caption{SCOPE architecture. The LM interleaves reasoning with PTZ actions (pan/tilt/zoom) and perception queries (counts, OCR, attributes). Results are fed back into context until a final answer is produced.}
  \Description{A block diagram of the SCOPE system: a central language model (LM) planner node connects to a PTZ camera controller (pan, tilt, zoom) and to a perception module that performs counting, OCR, and attribute queries. Arrows show results flowing back to the LM until a final answer is produced.}
  \label{fig:SCOPE-arch}
\end{figure}

Figure~\ref{fig:SCOPE-arch} illustrates the SCOPE architecture. The system adopts a decoupled design in which a compact SLM serves as a high-level planner, interleaving reasoning with camera control and perception queries, while visual understanding is delegated to lightweight VLMs exposed as callable tools. Rather than streaming image tokens into the dialogue—which can introduce hundreds to tens of thousands of tokens per frame and compound across turns ~\cite{moondream2,bai2025qwen2,wang2024mobile}—the planner invokes perception tools on demand and reasons over concise, task-aligned outputs. 

This separation reflects the current capabilities of open-source models: robust visual perception and reliable tool-based control remain difficult to achieve jointly within a single model at real-time edge latencies. By decoupling planning from perception, SCOPE achieves predictable latency, preserves modularity for independently upgrading planners and perception backbones, and improves safety and interpretability, enabling closed-loop PTZ control on local hardware.

Tool calls issued by the SLM are routed either to (i) the VLM, which returns grounded textual observations such as object counts, OCR results, or semantic attributes, or (ii) the PTZ camera API, which executes direct control commands. The PTZ path covers both viewpoint adjustments (pan, tilt, zoom) and direct outputs (e.g., screenshots), which can be appended to the agent’s context without invoking perception. All tool outputs are fed back into the planner’s context, enabling iterative reasoning and action until task completion.

\subsection{Agent Loop and Control}

At each step the agent queries the LM with the user request, history, and any tool observations. 
The model either terminates with a natural-language answer or issues tool calls; their outputs are appended into the dialogue state for the next step. 
This repeats until an answer is produced or a step cap $T_{\max}$ is reached.

Formally, SCOPE is a recurrent state machine with state \\ \mbox{$S_t = (U, H_t, O_t)$}, 
where $U$ denotes the fixed user query or goal, $H_t$ the cumulative dialogue and tool-call history, 
and $O_t$ the sequence of observed tool outputs. 
At each step $t$, the planner emits either a final answer or a structured tool-call sequence $\tau_t$, 
which yields the tool result $R_t$. 
The system then updates its internal memory as 
\mbox{$H_{t+1} = H_t \,\|\, [\tau_t, R_t]$} 
and observation log as 
\mbox{$O_{t+1} = O_t \,\|\, [R_t]$}. 
The process continues until a final answer is produced or the maximum reasoning horizon $T_{\max}$ is reached. This formulation captures SCOPE as a discrete-time decision process whose evolution depends on recurrent integration of actions and observations.

\subsection{Capabilities and Tool Interface}
\label{subsec:capandtools}

The SCOPE agent interacts with the camera and its environment exclusively through a fixed set of callable tools exposed to the SLM planner as OpenAI-compatible JSON schemas. Rather than emitting free-form actions, the planner selects from these tools and supplies structured arguments (e.g., numeric pan/tilt offsets or object descriptions), enforcing explicit, inspectable control over perception and actuation.

Table~\ref{tab:tool-interface} summarizes the full tool interface used by SCOPE. The interface comprises two categories of tools: direct camera control commands and perception tools backed by lightweight VLMs.

\begin{table}[h]
\centering
\caption{Tool interface exposed to the SLM planner in SCOPE.}
\label{tab:tool-interface}

\setlength{\tabcolsep}{4pt}
\renewcommand{\arraystretch}{1.05}

{\small
\begin{tabular}{@{}p{0.30\columnwidth}p{0.66\columnwidth}@{}}
\hline
\multicolumn{2}{l}{\textbf{Direct Camera Control}} \\
\hline
{\ttfamily ADJUST\_PTZ} & Applies relative pan/tilt/zoom adjustments (e.g., pan=$+10^\circ$, tilt=$+10^\circ$). \\
{\ttfamily GO\_TO\_PRESET} & Moves the camera to a named preset viewpoint. \\
{\ttfamily GO\_HOME} & Returns the camera to its home orientation. \\
{\ttfamily GET\_PRESETS} & Retrieves available preset viewpoints. \\
{\ttfamily TAKE\_IMAGE} & Captures the current camera frame. \\
\hline
\multicolumn{2}{l}{\textbf{Perception Tools}} \\
\hline
{\ttfamily COUNT\_OBJECTS} & Counts described objects in the current view or full panorama. \\
{\ttfamily QUERY\_ANSWER} & Answers a natural-language query about the current view or panorama. \\
{\ttfamily ZOOM\_TO\_OBJECT} & Localizes a target object and maps it to PTZ zoom/pan actions. \\
{\ttfamily TRACK\_OBJECT} & Repeatedly localizes a target and issues PTZ adjustments to keep it centered. \\
\hline
\end{tabular}
}
\end{table}



\setlength{\abovecaptionskip}{2pt}
\setlength{\belowcaptionskip}{2pt}
\renewcommand{\arraystretch}{1.25}
\setlength{\tabcolsep}{14pt}

\subsection{Real-World Setup} 
Development began initially on a physical system. An AXIS M5075-G PTZ camera was wrapped behind the same tool schema from Subsection~\ref{subsec:capandtools}, and a Streamlit interface exposed the camera livestream, a natural-language control panel, an annotated-frame gallery, and per-turn timing breakdowns.

This real deployment was essential for validating end-to-end feasibility and for uncovering practical failure modes. In particular, we observed that agent behavior could be destabilized by small prompt edits, with errors such as incorrect tool selection or inverted control actions (e.g., panning left when asked to pan right). Such sensitivity is well documented for SLMs, which can react strongly to minor changes in phrasing or prompt length~\cite{google2023prompt}. However, progress at this stage was necessarily qualitative—assessed by whether behavior “felt right” and lacked a systematic way to compare models, prompts, or configurations.

These limitations motivated the transition to a simulation-based evaluation framework that would support deterministic replay, controlled ablations, and scalable benchmarking.

\vspace{-6pt}
\subsection{Simulation Setup}


To enable repeatable evaluation and CI-style testing, we implemented the agent in simulation using Blender, an open-source 3D environment with a Python API. In Blender, the camera object can be entered directly and all intrinsic and extrinsic parameters are programmatically accessible. By fixing the camera position $(x,y,z)$, allowing yaw and pitch rotations, and controlling focal length to adjust field of view, we recreate the operational degrees of freedom of a PTZ camera.

\begin{figure}[t]
    \centering
    \includegraphics[width=0.65\linewidth]{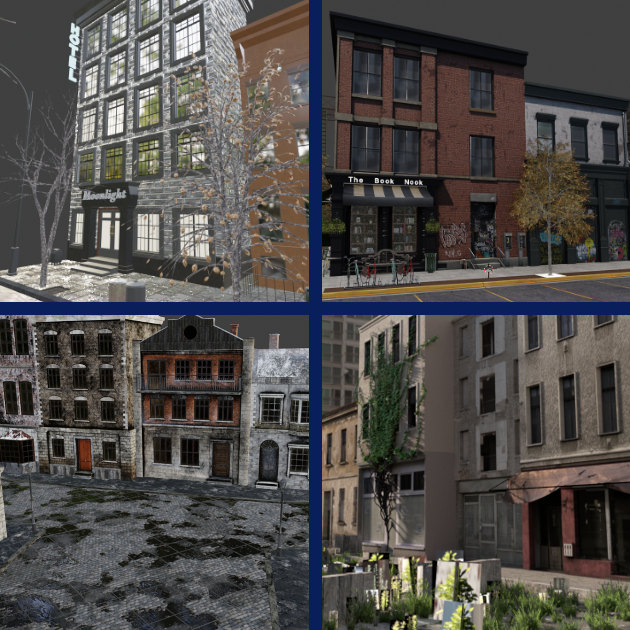}
    \caption{Representative Blender scenes used in simulation, showing diverse urban environments with signage, occlusion, and visual variation. Each scene includes multiple fixed camera presets at different locations and viewing angles.}
    \Description{A grid of four rendered Blender environments showing urban scenes with varied structure, signage, occlusion, and viewpoints.}
    \label{fig:example-worlds}
\end{figure}

All tools described in Subsection~\ref{subsec:capandtools} were re-implemented in this environment with identical names, parameters, and JSON schemas. From the SLM planner’s perspective, the interface is unchanged; only the backend differs (physical AXIS device versus Blender). This one-to-one correspondence enables direct comparability between simulation and real deployment. Representative simulated views are shown in Figure~\ref{fig:example-worlds}.

Some real-world queries require awareness beyond a single PTZ frame—for example, surveying an entire parking lot or reasoning about objects that may be outside the current view. Naïvely stitching images from different timeframes would introduce two major issues: (i) viewpoint overlap that could lead to double-counting or miscounts, and (ii) logical inconsistencies, since scenes can change between frames. 

To avoid these issues while keeping the evaluation focused, we implement a panorama tool that performs a single, deterministic sweep and stitches the resulting frames into a unified image representing the entire scene. This design allows us to evaluate two complementary capabilities: (i) the VLM’s ability to interpret stitched, multi-view imagery that departs from standard single-frame inputs, and (ii) the SLM’s ability to recognize when invoking a full-scene observation is necessary to answer a query.

The setup supports two complementary modes. First, one-off questions can be posed interactively inside Blender through a lightweight UI banner—useful for debugging or rapid testing. Second, scripted batch execution runs entire sets of questions sequentially across scenes, enabling large-scale evaluation or reinforcement learning experiments.

\vspace{-4pt}
\section{Benchmark Construction}
\subsection{Task Design and Coverage}
We construct a benchmark of 536 tasks to evaluate language-driven PTZ agents under realistic operating conditions. The benchmark spans action-oriented commands and question–answering tasks grounded in visual perception and tool use, with all tasks requiring explicit viewpoint changes. Typical examples include “sweep the scene and count all cars” (Counting), “zoom in on a storefront window and read the sign” (OCR), and “navigate to a predefined view and take a snapshot” (Multi-step Command). This design captures the core challenges of PTZ control beyond static image understanding. Table~\ref{tab:evalcats} reports the distribution of tasks across evaluation categories.

\subsection{Environment and Viewpoint Construction}

Because PTZ operation fundamentally relies on viewpoint control, we require environments in which a camera can meaningfully pan, tilt, and zoom rather than observe a static scene.
We therefore construct our benchmark using four publicly available Blender scenes representing diverse urban settings with signage, occlusion, clutter, and varying environmental conditions.

Within each scene, we define multiple fixed camera presets at distinct locations and viewing angles. 
These presets enable rapid, deterministic viewpoint changes while keeping the underlying environment fixed, allowing agents to survey a scene, zoom into regions of interest, or switch perspectives without introducing temporal inconsistency.
This setup mirrors real PTZ deployment, where cameras can have set predefined viewpoints. Representative views from these environments are shown in Figure~\ref{fig:example-worlds}.

\vspace{-6pt}

\subsection{Ground Truth Annotation}

To ground the tasks, we first created structured annotations for each camera preset. For every preset view, we documented observable entities and attributes (including counts and colors), spatial layout (e.g., left/right, near/far, and occlusions), OCR-relevant information, and how the scene would change under specific pan or tilt adjustments. We additionally recorded explicit absences (e.g., “no pedestrians or motorcycles in the parking lot”). These annotations serve as local ground truth: task prompts and expected answers were generated by an LLM conditioned on these annotations, ensuring that all tasks are anchored in verifiable observations rather than model-invented assumptions. Recording absences allowed an important extension--tasks whose correct outcome is the recognition that an object or attribute is not present.

\subsection{Task Construction and Dataset Refinement}

Task generation was seeded using a combination of the ground-truth annotations and a set of logged agent interaction traces from our physical PTZ deployment. These traces included full tool-call trajectories, intermediate observations, and operator feedback (e.g., thumbs up/down, some comments), which informed both representative user phrasing and expected behaviors. We additionally curated example task types for each evaluation category and provided guidance on a task distribution representative of realistic PTZ operation, spanning the categories summarized in Table~\ref{tab:evalcats}. This process produced an initial draft of 541 tasks.

Each task specifies a question–answer pair and, where applicable, includes structured expectations about agent behavior, such as required tool calls (or call order) and observation scope (current-view vs.\ full-scene). Together, these specifications capture both realistic user queries and challenging cases that stress perception, tool routing, and multi-step reasoning, enabling systematic analysis of where agents succeed and where they fail.

We then refined the dataset through an iterative validation process involving multiple representative SLM–VLM pairings and an LLM-as-Judge. We systematically audited low-success rate tasks and manually inspected tasks and agent trajectories to correct impossible or underspecified questions, remove duplicates, and align expected answers and tool-call expectations with the intended execution standard. In parallel, we refined the judge’s category-specific evaluation prompts to ensure that correctness judgments and error-mode attributions were consistent with our own assessments. Key adjustments included making open-ended prompts more permissive while retaining strict criteria for narrowly defined tasks.

Across three refinement passes, tasks and prompts were corrected, removed, or added as needed, the 536 tasks were finalized. The distribution of tasks can be seen in Table \ref{tab:evalcats}.

\section{Evaluation Methodology}\label{sec:eval_protocol}

This section defines the evaluation methodology used to assess PTZ agent performance. We introduce the scoring protocol for determining correctness and attributing failure modes, and then describe the model pairing strategy used to compare planners and perception backbones.

\subsection{Evaluation and Scoring Protocol}

To support structured error analysis, we define a set of mutually exclusive error categories shown in Table~\ref{tab:errormodes}. These categories capture failures arising from visual perception (e.g., misreading text or miscounting objects), reasoning errors, hallucinations, and tool-use mistakes such as missing calls, incorrect arguments, or improper scope selection.
\begingroup
\setlength{\textfloatsep}{4pt}   
\setlength{\intextsep}{4pt}      
\setlength{\floatsep}{4pt}       
\captionsetup[table]{skip=2pt}   

\begin{table}[ht]
\small
\centering
\setlength{\tabcolsep}{10pt}
\caption{Distribution of the 536 evaluation questions across categories. Tasks span visual perception (counting, descriptors, OCR, spatial reasoning) as well as control and reasoning skills (single calls, multi-step commands, and comparative queries).}
\resizebox{\columnwidth}{!}{%
  \begin{tabular}{l r | l r}
    \toprule
    \textbf{Evaluation Category} & \textbf{Count} &
    \textbf{Evaluation Category} & \textbf{Count} \\
    \midrule
    Counting & 95 & Single Call & 72 \\
    Descriptor & 89 & Multi-step Command & 57 \\
    Location / Spatial & 53 & Multi-step Reasoning & 54 \\
    OCR Identification & 54 & Comparative Relational & 62 \\
    \bottomrule
  \end{tabular}%
}
\label{tab:evalcats}
\end{table}

\endgroup
\enlargethispage{3\baselineskip}

\begin{table}[h]
\small
\centering
\caption{Error modes used in evaluation. When an answer is judged incorrect, the LM-as-Judge assigns one category to explain the failure.}
\Description{Table listing eight error modes and brief descriptions used to categorize incorrect PTZ agent executions.}
\label{tab:errormodes}

\setlength{\tabcolsep}{6pt}
\renewcommand{\arraystretch}{1.1}

\begin{tabular}{@{}p{0.29\columnwidth} p{0.66\columnwidth}@{}}
\toprule
\textbf{Error mode} & \textbf{Description} \\
\midrule
VLM -- Query       & Incorrect or incomplete visual read (OCR or attributes). \\
VLM -- Counting    & Miscounts objects or misses occluded items. \\
Reasoning          & Wrong conclusion despite correct tool outputs. \\
Hallucination      & Claims facts not grounded in tool evidence. \\
Lack of Tool Call  & Required tool never invoked. \\
Tool Args          & Incorrect arguments or targets (wrong preset/object). \\
Scope              & Wrong observation scope (\textit{current} vs.\ \textit{full}). \\
Tool Routing       & Wrong tool(s) chosen or required tool omitted. \\
\bottomrule
\end{tabular}
\end{table}

Table~\ref{tab:models} lists the SLM planners and VLMs evaluated in this study.
Model performance is assessed using an LLM-as-Judge framework that evaluates each agent execution holistically, rather than relying on deterministic string matching or rule-based checks. Such deterministic approaches break down for multi-step PTZ tasks, where multiple response phrasings, partial successes, or adaptive tool usage can all constitute valid outcomes. An LLM-based evaluator is therefore required to reliably assess both task completion and failure modes.

We use \mbox{gpt-oss-120B} as the sole LLM-as-Judge for all experiments~\cite{agarwal2025gpt}. For each agent execution, the judge is provided with the full trajectory—including tool calls, intermediate observations, and the final response—and produces a structured evaluation. Specifically, the judge returns (i) a binary correctness label and, for every incorrect execution, (ii) a single error attribution selected from the categories in Table~\ref{tab:errormodes}, along with (iii) a short natural-language rationale explaining the decision. Accuracy throughout the paper is therefore defined as the fraction of executions judged correct under this rubric, while error analysis is grounded in the explicit failure modes defined above.

We selected gpt-oss-120B because it is among the strongest publicly available open-weight models at the time of evaluation and was trained explicitly for agentic reasoning and tool-use tasks ~\cite{agarwal2025gpt}. Prior work and public benchmarks rank it competitively with leading proprietary judges: it performs comparably to GPT-4-class models on mixed-domain reasoning benchmarks ~\cite{chiang2024chatbot,shah2024stackeval} and has been used successfully for evaluating agent behaviors in recent systems work~\cite{agarwal2025gpt,hilton2025art}. Using a single, fixed judge and rubric across all model pairings ensures consistency in relative comparisons, even though the procedure is inherently non-deterministic.
\begin{table*}[ht!]
\small
\centering
\caption{Models evaluated in our study. Top block lists SLM planners, bottom block lists VLMs. For Mixture-of-Experts (MoE) models, ``Active Params'' refers to the number of parameters active per token; dense models show full size. Precision indicates the primary runtime format used in evaluation. Notes summarize release details and intended role (baseline, quantized variant, or efficiency-focused).}
\Description{Two-block table listing the language-model planners and vision-language models evaluated, including total size, active parameters, precision, and brief notes.}
\label{tab:models}

\setlength{\tabcolsep}{4pt}
\renewcommand{\arraystretch}{1.00}

\begin{tabular}{@{}p{0.24\textwidth} c c c p{0.45\textwidth}@{}}
\hline
\textbf{Language Model} & \textbf{Total Size} & \textbf{Active Params} & \textbf{Precision} & \textbf{Notes} \\
\hline
Qwen3-4B-Instruct-2507             & 4B  & 4B  & FP16 & Smaller dense model \cite{qwen3technicalreport} \\
Qwen3-4B-Instruct-2507-FP8         & 4B  & 4B  & FP8  & Smaller dense model, quantized \cite{qwen3technicalreport} \\
Qwen3-30B-A3B-Instruct-2507        & 30B & 3B  & FP16 & MoE, baseline \cite{qwen3technicalreport} \\
Qwen3-30B-A3B-Instruct-2507-FP8    & 30B & 3B  & FP8  & MoE, quantized baseline \cite{qwen3technicalreport} \\
Qwen3-32B-Instruct                 & 32B & 32B & FP16 & Dense baseline (older release) \cite{qwen3technicalreport} \\
Qwen3-Next-80B-A3B-Instruct        & 80B & 3B  & FP16 & Latest MoE baseline; deeper, faster thinking \cite{qwen3technicalreport} \\
Qwen3-Next-80B-A3B-Instruct-FP8    & 80B & 3B  & FP8  & Quantized variant of Next-80B \cite{qwen3technicalreport} \\
\multicolumn{5}{@{}l}{\textbf{Vision--Language Model}} \\
Moondream2-2025-06-21              & 2B & 2B & FP16 & Efficiency-focused VLM; ``superword'' tokenizer speeds answers by 20--40\% without accuracy drop \cite{moondream2} \\
Moondream2-4bit-2025-04-14         & 2B & 2B & INT4 & Quantization-aware trained; 42\% less memory, 0.6\% accuracy drop; 184 tok\,/\,s on RTX 3090 \cite{moondream2} \\
Moondream3-preview-2025-09-18      & 9B & 2B & FP16 & MoE VLM with 2B active experts; balances accuracy and efficiency \cite{moondream2} \\
Qwen2.5-VL-7B-Instruct             & 7B & 7B & FP16 & Full dense VLM; widely used \cite{bai2025qwen2} \\
\hline
\end{tabular}
\end{table*}

To improve reliability and alignment with human judgment, we adopted a human-in-the-loop evaluation process during benchmark construction and validation. This included manual inspection of judge rationales, iterative refinement of the scoring rubric, and review of full agent trajectories. These checks showed strong agreement between the judge’s decisions and our own assessments. In practice, the judge consistently penalized smaller SLMs for missing tool calls, incorrect routing, or hallucinated answers, while larger planners were more often credited for partial or correct executions—patterns that align with our qualitative observations during development. While we do not claim a formal external human study, this combination of a strong evaluator model, explicit rubric design, and extensive internal validation provides a robust and transparent basis for our evaluation.

\subsection{Model Selection and Pairing Strategy}

Our evaluation primarily pairs each SLM planner with the Moondream VLM family (Moondream2, Moondream2-4bit, and Moondream3), reflecting our deployment goal of edge-feasible perception. In our hardware setting, Moondream variants offered the most favorable latency–throughput trade-off while maintaining strong accuracy and clean JSON-based tool integration~\cite{moondream2}. Moondream2 reports 89.55\% accuracy on CountBench, while a comparable Qwen2.5-VL 72B model reports 93.6\%~\cite{moondream2,bai2025qwen2,paiss2023teaching}. Motivated by this gap, we include a targeted cross-family comparison pairing our largest planner, Qwen3-Next-80B-A3B, with Qwen2.5-VL-7B to examine whether benchmark trends and shared model lineage translate to downstream PTZ tasks. Due to the substantial GPU requirements of the VLM, this configuration was evaluated once as a controlled comparison.

\section{Results}\label{sec:results}
\begin{table*}[t]
\centering
\footnotesize
\caption[Evaluation-category accuracy by model]{Evaluation-category accuracy (\%) for each language model (rows), grouped by paired VLM family. Within each family, SLM planners are ordered by parameter scale (4B $\rightarrow$ 30B-A3B $\rightarrow$ 32B $\rightarrow$ 80B-A3B), with FP16 results listed before their quantized variants. Each cell reports accuracy (\%) for that category. ``Average'' denotes the unweighted mean accuracy across categories for each row. \textbf{Bold} indicates the highest score within a VLM family for a given category, while \underline{underlined} indicates the highest score across all families.}
\Description{Wide table reporting per-category accuracy for each planner, grouped by vision-language model family, with an average accuracy column.}
\label{tab:cat-metrics-by-model}

\setlength{\tabcolsep}{3.5pt}
\renewcommand{\arraystretch}{1.12}

\resizebox{\textwidth}{!}{%
\scriptsize
\begin{tabular}{l c c c c c c c c c}
\toprule
\textbf{Model} &
\shortstack[c]{Comparative\\Relational} &
Counting &
Descriptor &
\shortstack[c]{Location /\\Spatial} &
\shortstack[c]{Multi-step\\Command} &
\shortstack[c]{Multi-step\\Reasoning} &
\shortstack[c]{OCR\\Identification} &
\shortstack[c]{Single\\Call} &
Average \\
\midrule

\multicolumn{10}{l}{\textbf{Moondream2}}\\
Qwen3-4B & 27.4 & 68.4 & 38.2 & 52.8 & 89.5 & 64.8 & 38.9 & 97.2 & 59.7 \\
Qwen3-4B-FP8 & 27.4 & 69.5 & 40.4 & 56.6 & 89.5 & 61.1 & 40.7 & 97.2 & 60.3 \\
Qwen3-30B-A3B & 43.5 & 70.5 & 44.9 & 52.8 & 93.0 & 81.5 & 53.7 & \textbf{98.6} & 67.3 \\
Qwen3-30B-A3B-FP8 & \textbf{50.0} & 72.6 & \textbf{50.6} & 56.6 & 89.5 & 79.6 & \textbf{59.3} & \textbf{98.6} & 69.6 \\
Qwen3-32B & 41.9 & 70.5 & 46.1 & \textbf{64.2} & \textbf{96.5} & 72.2 & 53.7 & \textbf{98.6} & 68.0 \\
Qwen3-Next-80B-A3B & 46.8 & \underline{\textbf{76.8}} & 48.3 & 58.5 & \textbf{96.5} & \underline{\textbf{83.3}} & 55.6 & \textbf{98.6} & \textbf{70.6} \\

\addlinespace[4pt]
\multicolumn{10}{l}{\textbf{Moondream2-4bit}}\\
Qwen3-4B & 32.3 & 62.1 & 47.2 & 50.9 & 91.2 & 66.7 & 48.1 & \underline{\textbf{100.0}} & 62.3 \\
Qwen3-4B-FP8 & 27.4 & 62.1 & 48.3 & 52.8 & 89.5 & 66.7 & 46.3 & 97.2 & 61.3 \\
Qwen3-30B-A3B & 41.9 & 64.2 & 49.4 & 58.5 & \underline{\textbf{98.2}} & 72.2 & \textbf{51.9} & 98.6 & \textbf{66.9} \\
Qwen3-30B-A3B-FP8 & \textbf{50.0} & \textbf{65.3} & 55.1 & 50.9 & 93.0 & 72.2 & 40.7 & 98.6 & 65.7 \\
Qwen3-32B & 37.1 & 60.0 & \underline{\textbf{59.6}} & \textbf{64.2} & 87.7 & 68.5 & \textbf{51.9} & 98.6 & 65.9 \\
Qwen3-Next-80B-A3B & 36.5 & \textbf{65.3} & 52.8 & 58.5 & \underline{\textbf{98.2}} & \textbf{75.9} & 48.1 & 98.6 & 66.8 \\

\addlinespace[4pt]
\multicolumn{10}{l}{\textbf{Moondream3}}\\
Qwen3-4B & 32.3 & 68.4 & 33.7 & 47.2 & 87.7 & 55.6 & 66.7 & \underline{\textbf{100.0}} & 61.4 \\
Qwen3-4B-FP8 & 30.6 & 73.7 & 34.8 & 58.5 & 87.7 & 59.3 & 57.4 & 98.6 & 62.6 \\
Qwen3-30B-A3B & \underline{\textbf{56.5}} & \underline{\textbf{76.8}} & 37.1 & \underline{\textbf{66.0}} & \underline{\textbf{98.2}} & \textbf{79.6} & 77.8 & 98.6 & \underline{\textbf{73.8}} \\
Qwen3-30B-A3B-FP8 & 48.4 & 74.7 & 38.2 & 52.8 & 86.0 & 75.9 & 77.8 & 98.6 & 69.1 \\
Qwen3-32B & 45.2 & 68.4 & \textbf{39.3} & 52.8 & 89.5 & 72.2 & \underline{\textbf{83.3}} & \underline{\textbf{100.0}} & 68.8 \\
Qwen3-Next-80B-A3B & 45.2 & 72.6 & 34.8 & 47.2 & \underline{\textbf{98.2}} & 74.1 & \underline{\textbf{83.3}} & 98.6 & 69.3 \\

\addlinespace[4pt]
\multicolumn{10}{l}{\textbf{Qwen2.5-VL-7B}}\\
Qwen3-Next-80B-A3B & \textbf{33.9} & \textbf{52.6} & \textbf{52.8} & \underline{\textbf{66.0}} & \textbf{93.0} & \textbf{72.2} & \textbf{75.9} & \underline{\textbf{100.0}} & \textbf{68.3} \\

\bottomrule
\end{tabular}%
}
\end{table*}

\subsection{Where Errors Come From: A Decomposition}
Results in Table~\ref{tab:cat-metrics-by-model} reveal a structured error profile in language-driven PTZ agents, with distinct failure modes emerging from planner capacity and visual perception. Accuracy on visually demanding categories such as Counting (52.6--76.8\%) and OCR Identification (38.9--83.3\%) is substantially lower than on tool-centric categories like Single Call (97.2--100.0\%) and Multi-step Command (86.0--98.2\%), indicating that perception-heavy tasks are inherently more challenging. However, these aggregate trends mask important differences across planner capacity.

\begin{figure*}[ht]
  \centering
  \includegraphics[width=0.9\textwidth,alt={Grouped bar chart showing error counts by error category for multiple SLM–VLM combinations. Bars are color-coded by vision–language model family, with lighter to darker shades indicating increasing SLM size. Error categories include VLM query, counting, reasoning, hallucination, tool-call errors, scope errors, and tool routing.}]{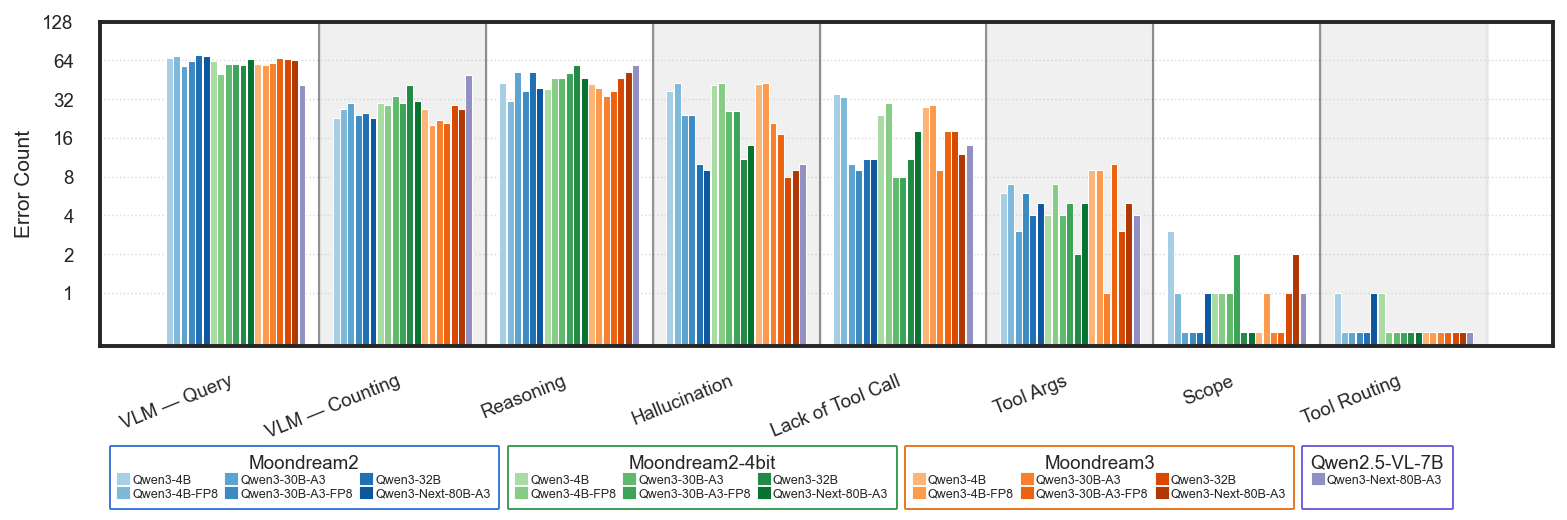}
  \caption{Error mode distribution across SLM–VLM combinations. Colors denote VLM selection, with shade intensity indicating increasing SLM size. Shaded bands and vertical dividers separate error categories.}
  \Description{Grouped bar chart illustrating the distribution of error counts across different error modes for all evaluated SLM–VLM combinations. Each group corresponds to an error category, such as VLM query errors, counting errors, reasoning errors, hallucinations, and tool-use mistakes. Bars are color-coded by vision–language model family, and darker shades represent larger or more capable SLM planners. The figure highlights how error profiles shift with planner capacity and perception backbone choice.}
  \label{fig:error-counts-by-mode}
\end{figure*}

Figure~\ref{fig:error-counts-by-mode} clarifies how failures are distributed across model pairings. For weaker SLM planners, errors frequently arise from hallucinations or improper tool usage, including argument-level mistakes such as issuing imprecise spatial references (e.g., confusing ``car'' with ``car to the right''). As planner capacity increases, these failure modes diminish markedly: stronger SLMs are more reliable, better grounded in tool outputs, and more consistent in invoking the correct tools over the correct observation scope. Tool-routing and scope-selection errors are rare for these planners, suggesting diminishing returns from further SLM scaling.

Among all combinations, failures are dominated by VLM querying and counting errors. For stronger SLM pairings, perception-related errors constitute the largest share of remaining failures, confirming that visual understanding becomes the primary limiter of end-to-end performance in the high-capacity planner regime. This shift highlights a transition from planner-limited to perception-limited behavior as SLM capacity increases.

Disentangling planner and perception effects further reinforces this asymmetry. Holding the VLM family fixed, changing only the SLM yields a substantial accuracy spread of 5.6--12.4\%, reflecting the importance of planner capacity for reducing hallucinations and reasoning failures. In contrast, the largest accuracy spread across VLM backbones is just 6.9\%. Notably, OCR and fine-grained counting tasks show the largest gains between Moondream2 and Moondream3, indicating that modular improvements to the vision backbone do translate into measurable gains.

To isolate the impact of perception more directly, we re-compute task accuracy by treating failures attributed to VLM counting and query errors as correct. Under this oracle setting, average accuracy increases from 66.2\% to 82.0\% (+15.8\%), indicating that perception failures account for 46.7\% of errors on average. Among the strongest SLM planners—where hallucination, routing, and reasoning errors are least frequent—perception constitutes an even larger fraction of remaining failures, confirming that the vision backbone ultimately sets the performance ceiling for PTZ agents.

\vspace{-6pt}

\subsection{Scope Sensitivity}
Query scope interacts strongly with perception difficulty. Current-view queries achieve 66.9\% average accuracy, compared to 58.3\% for full-scene queries. Given that scope-selection errors are rare, this gap likely reflects VLMs suffering from accuracy on panoramas, particularly on counting and text recognition tasks.

\begingroup
\setlength{\textfloatsep}{4pt}  
\setlength{\intextsep}{4pt}     
\setlength{\floatsep}{4pt}      
\begin{table}[ht]
\setlength{\textfloatsep}{4pt}   
\setlength{\intextsep}{4pt}
\setlength{\floatsep}{4pt}

\centering
\caption{Median inference latency (s) and mean tool calls across all SLM and VLM evaluations.}
\label{tab:latency-summary}
\renewcommand{\arraystretch}{1.05}
\setlength{\tabcolsep}{4pt}
\small
\begin{tabular*}{\columnwidth}{@{\extracolsep{\fill}} lcc@{}}
\toprule
\textbf{Model / Setting} & \textbf{Latency (s)} & \textbf{Tool Calls} \\
\midrule
\multicolumn{3}{l}{\textbf{Vision-Language Models}} \\
Qwen2.5-VL-7B & 5.46 & 1.34 \\
Moondream2 & 5.96 & 1.43 \\
Moondream2-4bit & 6.87 & 1.45 \\
Moondream3 & 9.18 & 1.50 \\
Current view $\rightarrow$ Panorama & 6.07 $\rightarrow$ 7.43 & — \\
\midrule
\multicolumn{3}{l}{\textbf{Small Language Models}} \\
Qwen3-4B & 1.94 & 1.44 \\
Qwen3-4B-FP8 & 0.96 & 1.41 \\
Qwen3-30B-A3B & 1.39 & 1.61 \\
Qwen3-30B-A3B-FP8 & 2.62 & 1.54 \\
Qwen3-32B & 3.00 & 1.51 \\
Qwen3-Next-80B-A3B & 1.42 & 1.63 \\
\bottomrule
\end{tabular*}
\end{table}
\endgroup

\subsection{Latency}
Table~\ref{tab:latency-summary} shows that end-to-end latency is dominated by perception rather than planning. VLM backbones vary widely in per-call inference cost, and expanding the observation scope to a full panorama increases latency by 22.4\% due to larger image inputs. In contrast, planner latency is comparatively stable across configurations: MoE planners achieve inference times close to those of small dense models despite much larger total capacity. Quantization reduces memory and can improve speed, but its latency impact is configuration-dependent in our setup, reflecting hardware, kernel, and serving-path effects. Overall, perception sets the dominant latency costs, while planner choice can improve accuracy without proportional inference overhead.

\vspace{-10pt}
\FloatBarrier
\vspace{-8pt}

\subsection{Architecture Selection}
Across VLM families, MoE SLMs achieve the highest accuracies while maintaining inference times comparable to dense planners. In practice, this indicates that MoE planners make more effective tool-selection and sequencing decisions without incurring a meaningful latency penalty. Quantization likewise preserves accuracy—performance differences are small and sometimes favor the quantized variant. Together, these results support MoE planners and quantization as practical deployment strategies for real-time PTZ agents. 

Consistent with these trends, the strongest SLM–VLM pairings combine capable MoE planners with the most accurate available vision backbones. The best-performing configuration, Moondream3 paired with Qwen3-30B-A3B, achieved 73.8\% task success.

\section{Limitations}
Because the benchmark is grounded in structured, worksheet-based prompts, queries are more constrained and less ambiguous than those encountered in real deployments. As a result, certain failure modes—such as incorrect tool sequencing, improper command scope, or recovery from ambiguous instructions—appear less frequently in aggregate metrics than they would in the field.

Additionally, latency measurements were likely skewed by differences in serving configurations rather than intrinsic model speed alone. Qwen SLM/VLM models were served using optimized vLLM-based infrastructure, whereas Moondream was evaluated in a non-optimized self-hosted setup. In some evaluation runs, multiple Moondream runs shared the same hardware, reducing effective throughput. When evaluated in isolation without resource contention, Moondream models consistently achieved lower, real-time per-call latency than the reported averages suggest.

Finally, we did not benchmark closed frontier models. While such systems may offer higher raw accuracy, their reliance on remote inference introduces network and queuing latency that is incompatible with real-time, language-guided PTZ control. Our evaluation therefore focuses on models that are realistically deployable in low-latency edge settings, and excludes black-box API models whose deployment characteristics fall outside the scope of this work.

\section{Discussion and Future Work}
Our findings show that compact MoE and quantized architectures provide the best accuracy–latency balance for real-time PTZ control. Dense 32B planners are impractically slow, while 30B MoE models—activating only 3B parameters—outperform smaller dense planners in reasoning. Quantization preserves accuracy but introduces A100-specific latency penalties due to kernel and serving-path limitations, underscoring that deployment performance depends on hardware and software support as much as model architecture. Scaling planners beyond 30B-A3B yields no clear accuracy gains at substantially higher memory cost, making 30B-A3B the most favorable deployment point.

While planner capacity is largely sufficient at this scale, residual planner-side gaps remain. Even strong SLMs occasionally fail to challenge ambiguous queries, reflect on prior tool outputs, or refine visual observations through narrower scope or second-pass queries. These issues—imprecise tool arguments, limited self-checking, and shallow use of conversational context—point to opportunities for improvement via prompt refinement, lightweight reflection, or RL that rewards corrective tool use.

Beyond these planner limitations, perception emerges as the dominant ceiling on end-to-end performance. Routing and sequencing errors are rare among MoE planners, but accuracy remains constrained by VLM failures in tasks such as OCR and fine-grained counting. This suggests future accuracy gains will hinge on stronger VLMs. Our latency results also highlight that for constrained subtasks (e.g., tracking), specialized detectors on lighter architectures such as YOLO may remain indispensable, pointing to hybrid stacks where the SLM orchestrates when and how to call them. Together, these findings validate our modular design: separating planning and perception makes real-time edge deployment practical today.

Future work will improve planner reliability via prompt refinement and reinforcement learning, expand SCOPE’s use of dynamic scene injection for tracking, and extend the framework to multi-camera, multi-agent deployments for cooperative, full-perimeter perception.

\balance
\bibliographystyle{ACM-Reference-Format}
\bibliography{ref}

@article{buynitsky2025camera,
  title={Camera Control at the Edge with Language Models for Scene Understanding},
  author={Buynitsky, Alexiy and Ehsani, Sina and Pallakonda, Bhanu and Mishra, Pragyana},
  journal={arXiv preprint arXiv:2505.06402},
  year={2025},
doi = {10.1109/ICCAR64901.2025.11073044}
}

@misc{qwen3technicalreport,
  title         = {{Qwen3} Technical Report},
  author        = {{Qwen Team}},
  year          = {2025},
  eprint        = {2505.09388},
  archivePrefix = {arXiv},
  primaryClass  = {cs.CL},
  url           = {https://arxiv.org/abs/2505.09388},
doi={10.48550/arXiv.2505.09388}
}

@article{liang2022code,
  title={Code as policies: Language model programs for embodied control},
  author={Liang, Jacky and Huang, Wenlong and Xia, Fei and Xu, Peng and Hausman, Karol and Ichter, Brian and Florence, Pete and Zeng, Andy},
  journal={arXiv preprint arXiv:2209.07753},
  year={2022},
doi={10.1109/ICRA48891.2023.10160591}
}

@article{vemprala2024chatgpt,
  title={ChatGPT for robotics: Design principles and model abilities},
  author={Vemprala, Sai H and Bonatti, Rogerio and Bucker, Arthur and Kapoor, Ashish},
  journal={IEEE Access},
  volume={12},
  pages={55682--55696},
  year={2024},
  publisher={IEEE},
doi={10.1109/ACCESS.2024.3387941}
}

@article{ahn2022can,
  title={Do as I can, not as I say: Grounding language in robotic affordances},
  author={Ahn, Michael and Brohan, Anthony and Brown, Noah and Chebotar, Yevgen and Cortes, Omar and David, Byron and Finn, Chelsea and Fu, Chuyuan and Gopalakrishnan, Keerthana and Hausman, Karol and others},
  journal={arXiv preprint arXiv:2204.01691},
  year={2022},
  doi={10.48550/arXiv.2204.01691}
}

@article{bai2025qwen2,
  title   = {{Qwen2.5-VL} Technical Report},
  author  = {Bai, Shuai and Chen, Keqin and Liu, Xuejing and Wang, Jialin and Ge, Wenbin and Song, Sibo and Dang, Kai and Wang, Peng and Wang, Shijie and Tang, Jun and others},
  journal = {arXiv preprint arXiv:2502.13923},
  year    = {2025},
  doi = {10.48550/arXiv.2502.13923},
}

@misc{hilton2025art,
  author = {Brad Hilton and Kyle Corbitt and David Corbitt and Saumya Gandhi and Angky William and Bohdan Kovalenskyi and Andie Jones},
  title = {ART: Agent Reinforcement Trainer},
  year = {2025},
  publisher = {GitHub},
  howpublished = {\url{https://github.com/openpipe/art}}
}

@inproceedings{deitke2025molmo,
  title     = {Molmo and Pixmo: Open Weights and Open Data for State-of-the-Art Vision--Language Models},
  author    = {Deitke, Matt and Clark, Christopher and Lee, Sangho and Tripathi, Rohun and Yang, Yue and Park, Jae Sung and Salehi, Mohammadreza and Muennighoff, Niklas and Lo, Kyle and Soldaini, Luca and others},
  booktitle = {Proceedings of the {IEEE}/{CVF} Conference on Computer Vision and Pattern Recognition ({CVPR})},
  pages     = {91--104},
  year      = {2025},
  publisher = {IEEE},
doi={10.48550/arXiv.2409.17146}
}

@inproceedings{patil2025bfcl,
  title        = {The Berkeley Function Calling Leaderboard (BFCL): From Tool Use to Agentic Evaluation of Large Language Models},
  author       = {Patil, Shishir G and Mao, Huanzhi and Yan, Fanjia and Ji, Charlie Cheng-Jie and Suresh, Vishnu and Stoica, Ion and Gonzalez, Joseph E.},
  booktitle    = {Proceedings of the 42nd International Conference on Machine Learning},
  series       = {Proceedings of Machine Learning Research},
  volume       = {267},
  year         = {2025},
  month        = {Jul},
  address      = {Vancouver, Canada},
  publisher    = {PMLR},
  url          = {https://openreview.net/forum?id=2GmDdhBdDk},
  note         = {ICML 2025 poster},
}

@article{hendrycks2020measuring,
  title={Measuring massive multitask language understanding},
  author={Hendrycks, Dan and Burns, Collin and Basart, Steven and Zou, Andy and Mazeika, Mantas and Song, Dawn and Steinhardt, Jacob},
  journal={arXiv preprint arXiv:2009.03300},
  year={2020},
  doi = {10.48550/arXiv.2009.03300},
}

@article{yue2024mmmu,
  title={MMMU-Pro: A More Robust Multi-discipline Multimodal Understanding Benchmark},
  author={Yue, Xiang and Zheng, Tianyu and Ni, Yuansheng and Wang, Yubo and Zhang, Kai and Tong, Shengbang and Sun, Yuxuan and Yu, Botao and Zhang, Ge and Sun, Huan and Su, Yu and Chen, Wenhu and Neubig, Graham},
  journal={arXiv preprint arXiv:2409.02813},
  year={2024},
doi={10.48550/arXiv.2409.02813}
}

@techreport{google2023prompt,
  title        = {Prompt Engineering},
  author       = {Boonstra, Lee},
  institution  = {Google Cloud},
  year         = {2024},
  month        = sep,
  type         = {Whitepaper},
  url          = {https://www.kaggle.com/whitepaper-prompt-engineering},
  urldate      = {2025-09-15}
}

@article{wang2024mobile,
  title={Mobile-agent: Autonomous multi-modal mobile device agent with visual perception},
  author={Wang, Junyang and Xu, Haiyang and Ye, Jiabo and Yan, Ming and Shen, Weizhou and Zhang, Ji and Huang, Fei and Sang, Jitao},
  journal={arXiv preprint arXiv:2401.16158},
  year={2024},
doi={10.48550/arXiv.2401.16158}
}

@inproceedings{shridhar2020alfred,
  title     = {ALFRED: A benchmark for interpreting grounded instructions for everyday tasks},
  author    = {Shridhar, Mohit and Thomason, Jesse and Gordon, Daniel and Bisk, Yonatan and Han, Winson and Mottaghi, Roozbeh and Zettlemoyer, Luke and Fox, Dieter},
  booktitle = {Proceedings of the IEEE/CVF Conference on Computer Vision and Pattern Recognition (CVPR)},
  pages     = {10740--10749},
  year      = {2020},
  publisher = {IEEE/CVF},
doi={10.1109/CVPR42600.2020.01075}
}

@inproceedings{das2018embodied,
  title     = {Embodied question answering},
  author    = {Das, Abhishek and Datta, Samyak and Gkioxari, Georgia and Lee, Stefan and Parikh, Devi and Batra, Dhruv},
  booktitle={Proceedings of the IEEE Conference on Computer Vision and Pattern Recognition},
  pages     = {1--10},
  year      = {2018},
  publisher = {IEEE},
  address   = {Salt Lake City, UT, USA},
doi = {10.1109/CVPR.2018.00008}
}

@misc{moondream2,
  title        = {Moondream: Lightweight Vision--Language Models ({Moondream2}, {Moondream2-4bit}, {Moondream3}-preview)},
  author       = {Karamcheti, Vikhyat},
  year         = {2025},
  howpublished = {\url{https://huggingface.co/moondream}},
  note         = {Includes Moondream2 (2025-06-21), Moondream2-4bit (2025-04-14), Moondream3-preview (2025-09-18)}
}

@misc{agarwal2025gpt,
  title         = {gpt-oss-120b \& gpt-oss-20b Model Card},
  author        = {Agarwal, Sandhini and Ahmad, Lama and Ai, Jason and Altman, Sam and Applebaum, Andy and Arbus, Edwin and Arora, Rahul K. and Bai, Yu and Baker, Bowen and Bao, Haiming and others},
  year          = {2025},
  eprint        = {2508.10925},
  archivePrefix = {arXiv},
  primaryClass  = {cs.CL},
  url           = {https://arxiv.org/abs/2508.10925},
  doi = {10.48550/arXiv.2508.10925}
}

@inproceedings{anderson2018vision,
  title     = {Vision-and-Language Navigation: Interpreting Visually-Grounded Navigation Instructions in Real Environments},
  author    = {Anderson, Peter and Wu, Qi and Teney, Damien and Bruce, Jake and Johnson, Mark and S{\"u}nderhauf, Niko and Reid, Ian and Gould, Stephen and Van Den Hengel, Anton},
  booktitle = {Proceedings of the {IEEE}/{CVF} Conference on Computer Vision and Pattern Recognition ({CVPR})},
  pages     = {3674--3683},
  year      = {2018},
  publisher = {IEEE},
doi={10.1109/CVPR.2018.00387}
}

@inproceedings{paiss2023teaching,
  title     = {Teaching {CLIP} to Count to Ten},
  author    = {Paiss, Roni and Ephrat, Ariel and Tov, Omer and Zada, Shiran and Mosseri, Inbar and Irani, Michal and Dekel, Tali},
  booktitle = {Proceedings of the {IEEE}/{CVF} International Conference on Computer Vision ({ICCV})},
  pages     = {3170--3180},
  year      = {2023},
  publisher = {IEEE},
  url       = {https://openaccess.thecvf.com/content/ICCV2023/html/Paiss_Teaching_CLIP_to_Count_to_Ten_ICCV_2023_paper.html},
doi={10.1109/ICCV51070.2023.00294}
}

@inproceedings{redmon2016yolo,
  title     = {You Only Look Once: Unified, Real-Time Object Detection},
  author    = {Redmon, Joseph and Divvala, Santosh and Girshick, Ross and Farhadi, Ali},
  booktitle = {Proceedings of the {IEEE} Conference on Computer Vision and Pattern Recognition ({CVPR})},
  pages     = {779--788},
  year      = {2016},
  publisher = {IEEE},
  doi = {10.1109/CVPR.2016.91}
}

@inproceedings{chiang2024chatbot,
  title={Chatbot arena: An open platform for evaluating llms by human preference},
  author={Chiang, Wei-Lin and Zheng, Lianmin and Sheng, Ying and Angelopoulos, Anastasios Nikolas and Li, Tianle and Li, Dacheng and Zhu, Banghua and Zhang, Hao and Jordan, Michael and Gonzalez, Joseph E and others},
  booktitle={Forty-first International Conference on Machine Learning},
  year={2024},
  doi={10.5555/3692070.3692401}
}

@inproceedings{shah2024stackeval,
  title={StackEval: Benchmarking LLMs in Coding Assistance},
  author={Shah, Nidhish and Genc, Zulkuf and Araci, Dogu},
  booktitle={Advances in Neural Information Processing Systems},
  volume={37},
  doi          = {10.52202/079017-1166},
  year={2024}
}

\end{document}